\documentclass[lettersize,journal]{IEEEtran}

\usepackage{cite}
\usepackage{amsmath,amssymb,amsfonts}
\usepackage{amsthm}
\usepackage{array}
\usepackage{graphicx}
\usepackage{booktabs}
\usepackage{multirow}
\usepackage{subcaption}
\usepackage{xcolor}
\usepackage{pifont}
\usepackage{textcomp}
\usepackage{stfloats}
\usepackage{url}
\usepackage[hidelinks]{hyperref}

\newcommand{\cmark}{\ding{51}} 
\newcommand{\xmark}{\ding{55}} 

\newtheorem{theorem}{Theorem}
\newtheorem{lemma}{Lemma}
\newtheorem{proposition}{Proposition}

\theoremstyle{remark}

\begin{document}

\title{Consist-Retinex: One-Step Noise-Emphasized Consistency Training Accelerates High-Quality Retinex Enhancement}

\author{Jian~Xu,
        Wei~Chen,
        Shigui~Li,
        Delu~Zeng,
        John~Paisley,
        and~Qibin~Zhao
\thanks{J. Xu and Q. Zhao are with the RIKEN Center for Advanced Intelligence Project (AIP), Tokyo, Japan (e-mail: jian.xu@riken.jp; qibin.zhao@riken.jp).}
\thanks{W. Chen, S. Li, and D. Zeng are with South China University of Technology, Guangzhou, China (e-mail: dlzeng@scut.edu.cn).}
\thanks{J. Paisley is with Columbia University, New York, NY, USA (e-mail: jwp2128@columbia.edu).}
}

\markboth{Preprint}%
{Xu \MakeLowercase{\textit{et al.}}: Consist-Retinex: One-Step Noise-Emphasized Consistency Training for Retinex Enhancement}

\maketitle

\begin{abstract}
Retinex-based low-light image enhancement benefits from separating reflectance and illumination, yet recent generative approaches often rely on iterative sampling and are difficult to deploy under strict latency budgets. Consistency models offer a natural route to one-step restoration, but direct adaptation to Retinex-factorized enhancement is unstable: one-step inference is evaluated at the high-noise endpoint, whereas standard training schedules provide little supervision there, and temporal self-consistency alone does not determine the correct conditional target. We propose Consist-Retinex, which first uses a Retinex Transformer Decomposition Network (TDN) to obtain paired reflectance and illumination maps, then trains two conditional consistency models with a Retinex-aware dual objective and adaptive noise-emphasized fixed-point sampling. The dual objective combines trajectory consistency with paired ground-truth component alignment, while the sampling rule concentrates supervision near the inference endpoint without discarding full-range noise coverage. We further provide an endpoint error bound, an anchoring-propagation result, and a high-noise sample-allocation analysis that explain why endpoint supervision and temporal consistency are complementary for one-step Retinex enhancement. Experiments show that the default Consist-Retinex improves step-efficient Retinex enhancement under a controlled backbone, while Consist-Retinex++ applies the same training strategy to the Diff-Retinex++-style backbone/augmentation setting and improves over the corresponding Diff-Retinex++ baseline. We therefore report the strongest numbers as a controlled same-architecture comparison, not as evidence that all gains come from consistency training alone.
\end{abstract}


\section{Introduction}
\label{sec:intro}

\IEEEPARstart{L}{ow-light} image enhancement is an important problem in computational photography, with applications ranging from mobile imaging to autonomous navigation. Retinex theory~\cite{land1977retinex} models an image as the product of reflectance and illumination, offering a useful factorization for brightening scenes while preserving structure. Modern Retinex-based networks and diffusion methods~\cite{wei2018deep,cai2023retinexformer,yi2023diff,yi2025diff} have improved restoration fidelity, but many high-quality generative pipelines still depend on iterative sampling or teacher trajectories, making direct one-step enhancement an attractive target.

Consistency models~\cite{song2023consistency} provide a promising alternative by learning mappings whose outputs remain self-consistent along probability-flow trajectories. They have been extended beyond unconditional generation to image restoration~\cite{gong2024ir} and fast super-resolution~\cite{xu2025fast}. However, Retinex-factorized low-light enhancement brings additional requirements. The model must predict two coupled normal-light components, reflectance and illumination, from the same degraded input; paired datasets provide direct targets; and one-step inference evaluates the network at the maximum noise level. As a result, a formulation designed only for unconditional self-consistency can be self-consistent but still anchored to the wrong conditional output.

This endpoint issue also exposes a sampling mismatch. Standard full-range log-uniform sampling rarely visits the interval used by one-step inference: with $\sigma_{\min}=0.002$ and $\sigma_{\max}=80$, only about $0.5\%$ of samples fall in $[0.95\sigma_{\max},\sigma_{\max}]$. For conditional enhancement, this high-noise region is where the low-light input must guide a direct jump to the normal-light component distribution. Undertraining this endpoint leads to unstable one-step predictions, while relying solely on low-noise or mid-noise consistency does not provide enough supervision for the actual inference regime.

We propose Consist-Retinex, a Retinex-oriented consistency training approach for one-step low-light enhancement. Given paired data $(I_l,I_n)$, we follow Diff-Retinex~\cite{yi2023diff} and use a TDN decomposition module to obtain low-light and normal-light components $(R_l,L_l)$ and $(R_n,L_n)$. Two conditional consistency models, $f_\theta^R$ and $f_\theta^L$, then map noisy normal-light component states to enhanced components conditioned on the corresponding low-light components. The final output is reconstructed as $\hat I=\hat R\odot\hat L$.

The core training design has two parts. First, a Retinex-aware dual-objective loss combines temporal consistency on a discrete Karras grid with paired ground-truth fixed-point alignment. Temporal consistency regularizes predictions along conditional trajectories, while fixed-point alignment selects the desired enhanced Retinex component rather than an arbitrary self-consistent value. Second, adaptive noise-emphasized sampling assigns most fixed-point supervision to $\sigma\approx\sigma_{\max}$ while preserving sparse full-range coverage, directly matching the one-step inference endpoint.

We also analyze this design from a consistency-model perspective. The endpoint error bound shows that one-step Retinex inference is controlled by the high-noise fixed-point error plus an explicit inference-prior mismatch term. The anchoring proposition shows how temporal consistency propagates a correct high-noise target along the trajectory. The sample-allocation result explains why increasing the high-noise sampling probability reduces the number of fixed-point samples needed to cover the endpoint regime. These results connect the dual objective and adaptive noise-emphasized training to the empirical ablations.

Our contributions are:
\begin{itemize}
    \item We formulate dual-objective conditional consistency training for TDN-decomposed Retinex components, combining temporal self-consistency with paired fixed-point alignment.
    \item We develop adaptive noise-emphasized fixed-point sampling that focuses supervision near the one-step inference endpoint, and validate its effect through component, threshold, and loss-weight ablations.
    \item We provide endpoint-oriented theoretical analysis, including a one-step error bound paired with an explicit EDM-preconditioning contraction lemma that quantifies the prior-mismatch term, a degeneracy result showing that consistency-only training admits an infinite-dimensional null space and is therefore insufficient for conditional Retinex enhancement, an anchoring-propagation result, and high-noise sample-allocation and concentration arguments for the proposed training strategy.
    \item We emphasize controlled comparisons: the default Consist-Retinex evaluates the proposed training strategy under the base Retinex backbone, while Consist-Retinex++ uses the Diff-Retinex++-style backbone/augmentation setting and improves over the corresponding Diff-Retinex++ baseline with one-step inference.
\end{itemize}
\section{Related Work}
\label{sec:related}

\subsection{Retinex-Based Low-Light Image Enhancement}

\textbf{Classical Retinex Theory.}
Retinex theory~\cite{land1977retinex}, modeling the human visual system's color perception, decomposes observed images into reflectance (intrinsic object properties) and illumination (lighting conditions) as $I = R \odot L$, where $\odot$ denotes element-wise multiplication. Early methods like SSR~\cite{jobson1997properties} and MSR~\cite{jobson1997multiscale} employed Gaussian filtering, while LIME~\cite{guo2016lime} initialized illumination via channel-wise maxima with structure-aware refinement. JED~\cite{ren2018joint} combined sequential decomposition with gamma correction for joint enhancement and denoising. However, hand-crafted priors in these traditional approaches suffer from limited adaptability and computational inefficiency.

\textbf{Deep Retinex Methods.}
The integration of deep learning with Retinex theory marked a paradigm shift. RetinexNet~\cite{wei2018deep} pioneered end-to-end decomposition using CNNs, followed by BM3D~\cite{dabov2006image} for reflectance denoising. KinD~\cite{zhang2019kindling} and KinD++~\cite{zhang2021beyond} introduced learnable decomposition--restoration frameworks with dedicated denoising modules. URetinex-Net~\cite{wu2022uretinex} unfolded optimization into deep networks via alternating half-quadratic splitting, achieving interpretability while learning implicit priors. Recent works like Retinexformer~\cite{cai2023retinexformer} leverage Transformers for decomposition. Despite progress, these methods face two critical limitations: (1) \textit{two-stage decomposition-then-enhancement paradigms} accumulate errors and deviate from holistic human visual perception, and (2) \textit{dedicated denoising on reflectance} often sacrifices fine details. While these feed-forward Retinex models are themselves efficient at inference, the recent generative Retinex approaches discussed next often require tens to thousands of iterative sampling steps, which constrains real-time applicability.

\subsection{Diffusion Models for Low-Light Enhancement}

\textbf{Diffusion Models Fundamentals.}
Denoising Diffusion Probabilistic Models (DDPM)~\cite{ho2020denoising} learn data distributions by progressively adding Gaussian noise in a forward process, then reversing this process via learned denoising steps. The probability flow ODE~\cite{song2021scorebased} provides a deterministic sampling pathway, enabling faster generation via advanced ODE solvers~\cite{lu2022dpm}. DDPMs have achieved strong image-generation quality~\cite{dhariwal2021diffusion,rombach2022high} through iterative refinement.

\textbf{Diffusion-Based LLIE Methods.}
Recent works have applied diffusion models to low-light enhancement. Diff-Retinex~\cite{yi2023diff} combined Retinex decomposition with separate diffusion processes for illumination and reflectance components, improving texture restoration through iterative sampling. Its extension Diff-Retinex++~\cite{yi2025diff} integrated Retinex theory as a plug-and-play supervision attention module within a unified diffusion framework, coupled with mixture-of-experts for multi-scenario adaptation. CLE Diffusion~\cite{yin2023cle} addressed color shift via controllable enhancement, while PyDiff~\cite{zhou2023pyramid} employed pyramid structures with global color correction. GSAD~\cite{hou2023global} incorporated global structure awareness. However, many diffusion-based LLIE methods still require long sequential denoising chains (typically 1000 steps for Diff-Retinex/++), which limits deployment in latency-sensitive settings despite their strong quality.

\subsection{Consistency Models for Fast Generation}

\textbf{Consistency Model Foundations.}
Song et al.~\cite{song2023consistency} recently introduced consistency models to enable one-step generation from diffusion models. By enforcing self-consistency---that any point on a probability flow ODE trajectory maps to the same origin---consistency models learn direct noise-to-data mappings. Two training paradigms exist: (1) \textit{consistency distillation} leverages pre-trained diffusion models to generate trajectory pairs, distilling multi-step knowledge into single-step inference, and (2) \textit{consistency training} trains models from scratch using temporal consistency across adjacent time steps. Beyond unconditional synthesis, IR-CM~\cite{gong2024ir} studies consistency models for general-purpose image restoration, and consistency rectified flow~\cite{xu2025fast} accelerates image super-resolution. Our setting differs from IR-CM in three ways: we operate on Retinex-decomposed reflectance/illumination components rather than a single generic image-restoration mapping; we use a dual objective that combines temporal consistency with paired component-level fixed-point alignment; and we introduce noise-emphasized endpoint sampling for one-step Retinex inference. These ingredients are specific to conditional low-light enhancement and are evaluated through the ablations in Tables~\ref{tab:ablation}, S3, and~S4.

\textbf{The Conditional Generation Gap.}
Despite their success in unconditional settings and recent restoration extensions, consistency models remain less developed for Retinex-factorized low-light enhancement. Song et al.~\cite{song2023consistency} provide neither loss formulations nor sampling strategies adapted to conditioning. Standard unconditional schedules are designed for transforming noise into samples, but they do not explicitly emphasize the high-noise endpoint used by one-step conditional inference. In conditional enhancement, large-noise regions ($\sigma \approx \sigma_{\max}$) are where degraded inputs must be integrated with target distributions---precisely the regime under-emphasized by full-range fixed-point sampling. Moreover, existing consistency objectives lack Retinex-specific ground-truth alignment terms applicable when paired data $(x_{\text{low}}, x_{\text{high}})$ are available. This mismatch between unconditional design and Retinex conditional requirements motivates our work.
\section{Method}
\label{sec:method}

\subsection{Problem Setup and Preliminaries}
\label{subsec:preliminaries}

\noindent\textbf{TDN-based Retinex decomposition.}
Retinex theory~\cite{land1977retinex} decomposes images as $I=R\odot L$, where $R$ is reflectance and $L$ is illumination. Following Diff-Retinex~\cite{yi2023diff}, we use a Retinex Transformer Decomposition Network (TDN) $D_\phi$ to obtain paired low-/normal-light components:
\begin{equation}
    (R_l,L_l)=D_\phi(I_l), \qquad (R_n,L_n)=D_\phi(I_n).
    \label{eq:retinex_targets}
\end{equation}
TDN is trained as the first stage with standard Retinex reconstruction, cross-reconstruction, reflectance-consistency, and illumination-smoothness losses. The subsequent consistency models are trained on the resulting component maps, with $(R_n,L_n)$ serving as learned normal-light component targets.

\noindent\textbf{Diffusion and consistency models.}
Following EDM~\cite{karras2022elucidating}, we use a continuous noise scale $\sigma\in[0.002,80]$. Consistency models~\cite{song2023consistency} learn a direct map whose outputs remain self-consistent along a trajectory, i.e., $f(x_t,t)=f(x_{t'},t')$ for states from the same sample. We adopt EDM preconditioning:
\begin{equation}
    f_\theta(x, t) = c_{\text{skip}}(t)\,x + c_{\text{out}}(t)\,F_\theta(x,t),
\end{equation}
where $c_{\text{skip}}(t)$ and $c_{\text{out}}(t)$ follow the standard EDM form with $\sigma_{\text{data}}=0.5$.
\subsection{Conditional Consistency for Retinex Enhancement}
\label{subsec:conditional_formulation}

\begin{figure*}[t]
    \centering
    \includegraphics[width=0.9\linewidth]{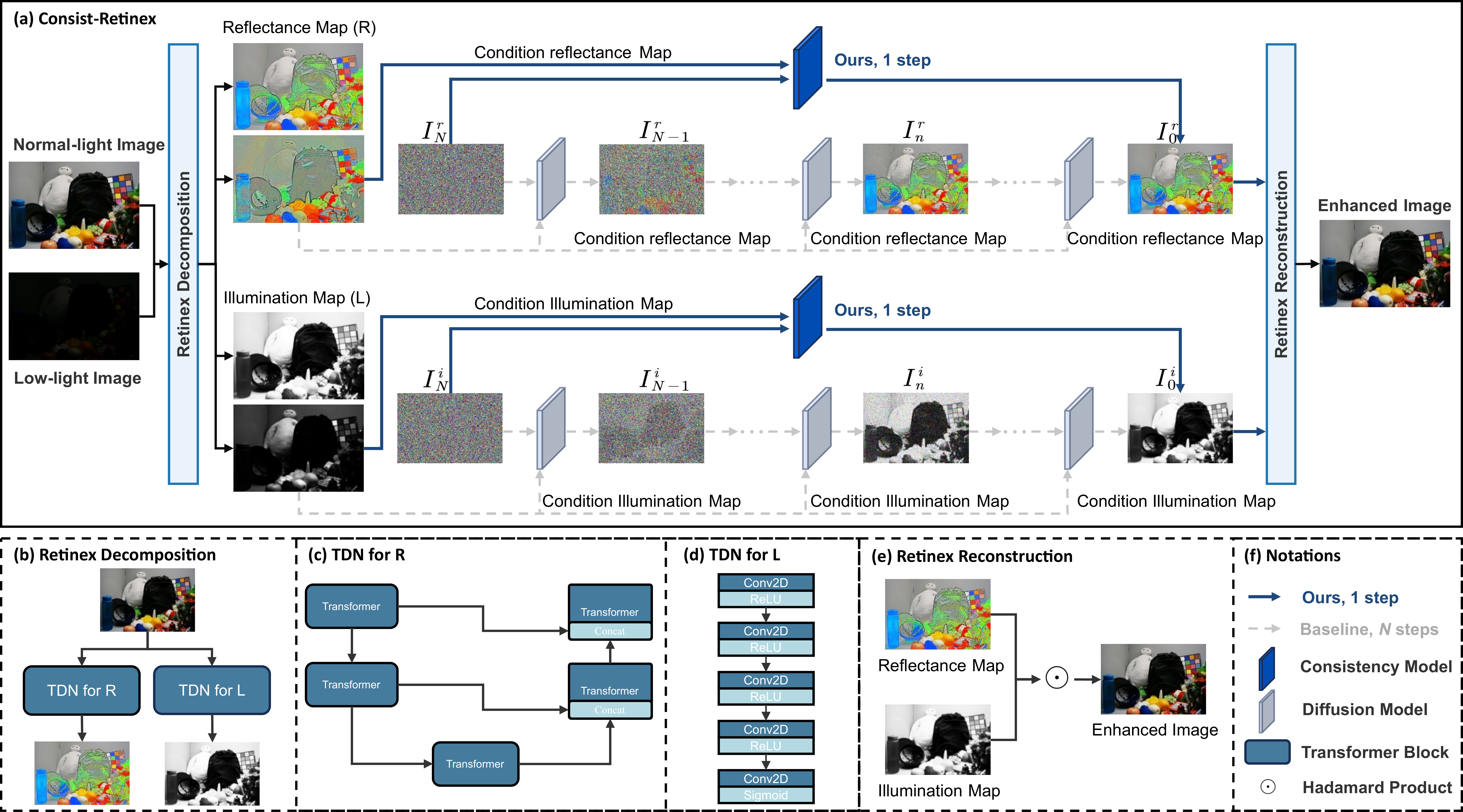}
    \caption{Overview of Consist-Retinex framework. Given a low-light image $I_l$, TDN decomposes it into low-light Retinex components, and two conditional consistency models $f_\theta^R$ and $f_\theta^L$ directly map noisy normal-light reflectance and illumination states to enhanced component outputs in one step. The final result is reconstructed via element-wise multiplication.}
    \label{fig:architecture}
\end{figure*}

\noindent\textbf{Framework overview.}
Figure~\ref{fig:architecture} illustrates our Consist-Retinex framework. Given a low-light image $I_l$, TDN first produces $(R_l,L_l)=D_\phi(I_l)$. We then learn two conditional consistency models:
\begin{align}
    \hat{R}_n &= f_\theta^R(R_n + \sigma \epsilon_R, \sigma \mid R_l), \label{eq:R_model}\\
    \hat{L}_n &= f_\theta^L(L_n + \sigma \epsilon_L, \sigma \mid L_l), \label{eq:L_model}
\end{align}
where $\epsilon_R, \epsilon_L \sim \mathcal{N}(0,\mathbf{I})$. The enhanced output is $\hat{I}_n = \hat{R}_n \odot \hat{L}_n$. At inference, we first compute $(R_l,L_l)=D_\phi(I_l)$ and then perform one-step generation from maximum noise: $\hat{R}_n = f_\theta^R(\sigma_{\max} \epsilon_R, \sigma_{\max} \mid R_l)$ and $\hat{L}_n = f_\theta^L(\sigma_{\max} \epsilon_L, \sigma_{\max} \mid L_l)$, without iterative denoising. Fig.~S1 of the supplementary material visualizes the consistency mapping used by these component models.

\noindent\textbf{Training dynamics mismatch.}
Our one-step conditional inference operates at $\sigma = \sigma_{\max}$, where the low-light TDN components must integrate with target distributions $(R_n,L_n)$. If fixed-point supervision is sampled from the full log-uniform range $\log\mathcal{U}(\sigma_{\min},\sigma_{\max})$, the probability of hitting the endpoint interval $[0.95\sigma_{\max},\sigma_{\max}]$ is only $\log(1/0.95)/\log(\sigma_{\max}/\sigma_{\min})\approx0.5\%$. This leaves the one-step inference regime under-supervised. Moreover, existing objectives lack explicit ground-truth alignment to paired TDN component targets.

\subsection{Noise-Emphasized Adaptive Sampling}
\label{subsec:adaptive_sampling}

\begin{figure*}[!htbp]
    \centering
    \includegraphics[width=0.52\linewidth]{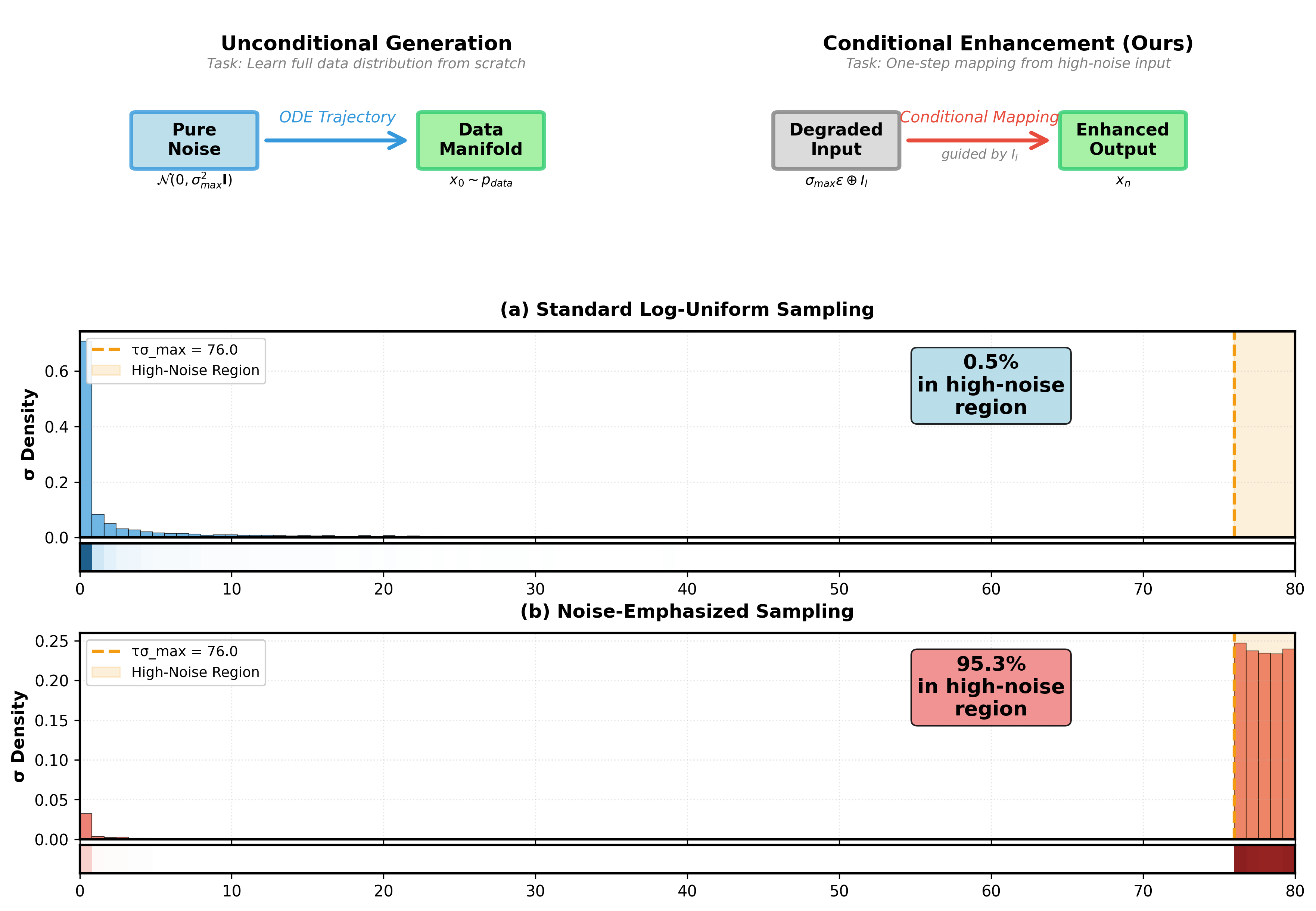}
  \caption{Task-driven sampling strategy comparison.
\textit{Top}: Unconditional generation learns from pure noise, while our conditional enhancement performs one-step mapping from concatenated noise and degraded inputs.
\textit{Middle}: Full-range log-uniform sampling provides little probability mass in the narrow high-noise endpoint interval.
\textit{Bottom}: Our noise-emphasized sampling focuses on high-noise regions critical for one-step conditional inference.}

    \label{fig:sampling}
\end{figure*}

We use two distinct noise schedules. Temporal consistency is defined on a discrete Karras grid in Sec.~\ref{subsec:dual_objective}, while the fixed-point alignment term uses the following continuous noise-emphasized mixture (Fig.~\ref{fig:sampling}):
\begin{equation}
\label{eq:bimodal_sampling}
    \sigma_{\text{rand}} \sim
    \begin{cases}
        \log \mathcal{U}(0.95\sigma_{\max},\,\sigma_{\max}), & \text{w.p. } 0.95, \\
        \log \mathcal{U}(\sigma_{\min},\,\sigma_{\max}), & \text{w.p. } 0.05.
    \end{cases}
\end{equation}
This allocates most training to $\sigma \in [76, 80]$ where one-step inference operates, while retaining sparse full-spectrum coverage for generalization. Empirically, low-noise loss converges rapidly (few thousand iterations), while high-noise mapping remains challenging without targeted emphasis.

\subsection{Dual-Objective Consistency Loss}
\label{subsec:dual_objective}

We combine two complementary terms with different sampling strategies: temporal consistency for trajectory coherence on a Karras grid and ground-truth alignment for conditional mapping under noise-emphasized fixed-point sampling. For a generic Retinex component $Z\in\{R,L\}$, let $x_0=Z_n$ and let $C_l^Z$ denote the corresponding low-light TDN condition ($C_l^R=R_l$, $C_l^L=L_l$).

\noindent\textbf{Temporal consistency.}
Following~\cite{song2023consistency,karras2022elucidating}, we enforce self-consistency on a discrete Karras noise grid
\begin{equation}
    \sigma_i =
    \left[
    \sigma_{\max}^{1/\rho}
    + \frac{i-1}{N-1}
    \left(\sigma_{\min}^{1/\rho}-\sigma_{\max}^{1/\rho}\right)
    \right]^\rho,
    \quad i=1,\ldots,N,
    \label{eq:karras_grid}
\end{equation}
where $\sigma_1=\sigma_{\max}$, $\sigma_N=\sigma_{\min}$, $N=10$, and $\rho=7$. We sample an index $n$ uniformly and an index gap $g\sim\mathcal{U}\{1,5\}$ with $n+g\leq N$. Given $x_{\sigma_n}=x_0+\sigma_n\epsilon$, the EMA model predicts $\hat{x}^{\text{target}}_0=\text{sg}(f_{\theta^-}(x_{\sigma_n},\sigma_n\mid C_l^Z))$. Euler approximation gives the lower-noise state
\begin{equation}
    \tilde{x}_{\sigma_{n+g}}
    =
    x_{\sigma_n}
    +(\sigma_{n+g}-\sigma_n)\cdot
    \frac{x_{\sigma_n}-\hat{x}^{\text{target}}_0}{\sigma_n}.
\end{equation}
The loss is:

\begin{equation}
\begin{aligned}
    \mathcal{L}_{\text{consist}} = \mathbb{E}_{\epsilon, n, g}\Big[ w(\sigma_n)\,\big\|
    & f_\theta(x_{\sigma_n}, \sigma_n \mid C_l^Z) \\
    & - \text{sg}\big(f_\theta(\tilde{x}_{\sigma_{n+g}}, \sigma_{n+g} \mid C_l^Z)\big)\big\|_2^2\Big],
    \end{aligned}
    \end{equation}
with SNR-based weight $w(\sigma_n) = (\sigma_{\text{data}}/\sigma_n)^2/[1 + (\sigma_{\text{data}}/\sigma_n)^2]$. Target network updates use the training iteration $s$, not the index gap: $\theta^- \leftarrow \mu_s \theta^- + (1-\mu_s)\theta$ with $\mu_s = \min(0.9999, (1+s)/(10+s))$.

\noindent\textbf{Ground-truth alignment.}
We anchor predictions to enhanced components using noise-emphasized sampling:
\begin{equation}
    \mathcal{L}_{\text{fixed}} = \mathbb{E}_{\epsilon,\sigma_{\text{rand}}}\left[\left\|f_\theta(x_0 + \sigma_{\text{rand}} \epsilon, \sigma_{\text{rand}} \mid C_l^Z) - x_0\right\|_2^2\right],
\end{equation}
where $x_0$ denotes either TDN-derived target $R_n$ or $L_n$ from Eq.~\ref{eq:retinex_targets}, and $\sigma_{\text{rand}}$ follows Eq.~\ref{eq:bimodal_sampling}. This concentrates supervision at $\sigma \approx \sigma_{\max}$ where inference occurs.

\noindent\textbf{Combined objective.}
The final loss is:
\begin{equation}
    \mathcal{L}_{\text{total}} = \lambda_{\text{consist}}\,\mathcal{L}_{\text{consist}} + \lambda_{\text{fixed}}\,\mathcal{L}_{\text{fixed}},
\end{equation}
with $\lambda_{\text{consist}} = 1.0$ and $\lambda_{\text{fixed}} = 0.3$. This dual-objective, dual-sampling design yields stable training: Consist-Retinex converges in 200K iterations (batch size 8) vs. 800K iterations (batch size 16) for diffusion baselines~\cite{yi2023diff,yi2025diff}.

\noindent\textbf{Empirical hyperparameter choice.}
We set $\lambda_{\text{consist}}=1.0$ as the reference consistency scale and tune $\lambda_{\text{fixed}}$ relative to it. The component ablation in Table~\ref{tab:ablation} shows that both temporal consistency and fixed-point alignment are necessary: GT-only regression is clearly weaker than consistency-guided variants, while removing $\mathcal{L}_{\text{fixed}}$ collapses performance. The detailed sweep in supplementary Table~S4 indicates that $\lambda_{\text{fixed}}=0.3$ gives the best balance between endpoint anchoring and trajectory coherence: $\lambda_{\text{fixed}}=0$ collapses performance, whereas larger values (e.g., $0.5$ or $1.0$) slightly degrade results by over-emphasizing direct regression. Because $\mathcal{L}_{\text{fixed}}$ is evaluated under noise-emphasized sampling, its effectiveness also depends on the sampling hyperparameters. Supplementary Table~S3 shows that $\tau=0.95$ (equivalently emphasizing $\sigma\in[0.95\sigma_{\max},\sigma_{\max}]$ with $p_{\text{large}}=0.95$ in Eq.~\ref{eq:bimodal_sampling}) provides the best trade-off between focusing on the one-step inference regime and retaining enough noise diversity for stable optimization.

\subsection{One-Step Inference}
\label{subsec:inference}
\textbf{One-step generation (default).} At test time, given a low-light image $I_l$, Consist-Retinex performs:
\begin{enumerate}
    \item \textbf{TDN decomposition:} compute $(R_l,L_l)=D_\phi(I_l)$.

    \item \textbf{Noise sampling:} draw $\epsilon_R, \epsilon_L \sim \mathcal{N}(0, I)$.
    
    \item \textbf{One-step enhancement:}
    \begin{align}
    \hat{R}_n &= f^\text{R}_\theta(\sigma_{\max}\epsilon_R, \sigma_{\max} \mid R_l), \\
    \hat{L}_n &= f^\text{L}_\theta(\sigma_{\max}\epsilon_L, \sigma_{\max} \mid L_l).
    \end{align}
    
    \item \textbf{Reconstruction:}
    \begin{equation}
    \hat{I}_n = \text{clip}(\hat{R}_n \odot \hat{L}_n, -1, 1).
    \end{equation}
\end{enumerate}
After TDN decomposition, this requires a single consistency-model forward pass per component. An optional light-weight multi-step refinement variant, which trades roughly $4\times$ the compute for about $+0.5$~dB on VE-LOL-L, is described in Section~S1 of the supplementary material.

\subsection{Architecture}
\label{subsec:architecture}

\noindent\textbf{Input conditioning.}
We concatenate the corresponding low-light TDN component and scaled noisy target component: $\text{Input} = [Z_l,\; c_{\text{in}}(\sigma)\cdot(Z_n + \sigma \epsilon)]$, where $Z\in\{R,L\}$ and $c_{\text{in}}(\sigma) = 1/\sqrt{\sigma^2 + \sigma_{\text{data}}^2}$. This yields $[B,6,H,W]$ for reflectance and $[B,2,H,W]$ for illumination.

\noindent\textbf{Time embedding.}
We embed $\sigma$ via sinusoidal Fourier features (256 bands) followed by MLP, injected through Adaptive Group Normalization~\cite{dhariwal2021diffusion}:
\begin{equation}
    \text{GN}(h)\,(1 + \gamma(t_{\text{emb}})) + \beta(t_{\text{emb}}),
\end{equation}
where $\gamma(\cdot)$ and $\beta(\cdot)$ are learned affine transforms.

\noindent\textbf{Model configurations.}
The reflectance consistency model $f^\text{R}_\theta$ adopts a hybrid architecture combining Restormer~\cite{zamir2022restormer} and U-Net components, which has demonstrated superior performance for reflectance restoration in Diff-Retinex~\cite{yi2023diff}. The Restormer backbone employs multi-Dconv head transposed attention (MDTA) blocks and gated-Dconv feed-forward networks (GDFN) that efficiently capture long-range dependencies for texture detail preservation, while the U-Net structure provides hierarchical feature extraction. The model configuration uses base channel width 64, channel multipliers $\{1, 2, 4, 6\}$, with 1-1-2 TransformerBlocks across three encoder levels, 2 bottleneck blocks at the deepest level, and 2 refinement blocks ($\sim$45M parameters). The illumination model $f_\theta^L$ uses a compact U-Net with base width 32, multipliers $\{1,2,2,4\}$, and attention at resolution 16 ($\sim$12M parameters).

\begin{table*}[ht]
\centering
\caption{Quantitative comparison on LOL and VE-LOL-L. Bold indicates the best result. For our rows, all metrics report mean$\pm$std over six stochastic one-step samplings. Consist-Retinex++ is the same-training-strategy comparison under the Diff-Retinex++-style backbone/augmentation setting; see Section~\ref{subsubsec:plus_variant}.}
\label{tab:lol_velol_results}
\resizebox{\textwidth}{!}{
\setlength{\tabcolsep}{3pt}
\begin{tabular}{@{}l|ccccc|ccccc@{}}
\hline
\multirow{2}{*}{Method} & \multicolumn{5}{c|}{LOL} & \multicolumn{5}{c}{VE-LOL-L} \\ 
\cline{2-11}
 & PSNR & SSIM & LPIPS & FID & MAE & PSNR & SSIM & LPIPS & FID & MAE \\ 
\hline
NPE~\cite{wang2013naturalness} & 16.97 & 0.484 & 0.400 & 118.421 & 0.142 & 17.333 & 0.464 & 0.396 & 101.875 & 0.147 \\
LIME~\cite{guo2016lime} & 16.554 & 0.429 & 0.405 & 114.003 & 0.123 & 15.105 & 0.402 & 0.426 & 98.905 & 0.145 \\
RUAS~\cite{liu2021retinex} & 16.405 & 0.500 & 0.270 & 112.354 & 0.153 & 15.326 & 0.488 & 0.310 & 100.078 & 0.162 \\
SCI~\cite{ma2022toward} & 14.784 & 0.525 & 0.333 & 110.285 & 0.135 & 17.304 & 0.540 & 0.307 & 99.231 & 0.151 \\
Zero-DCE~\cite{guo2020zero} & 19.524 & 0.703 & 0.330 & 106.629 & 0.106 & 18.059 & 0.574 & 0.313 & 91.939 & 0.131 \\
SGZ~\cite{zheng2022semantic} & 14.546 & 0.436 & 0.353 & 115.904 & 0.143 & 16.992 & 0.359 & 0.338 & 100.875 & 0.146 \\
KinD~\cite{zhang2019kindling} & 17.648 & 0.775 & 0.175 & 78.586 & 0.123 & 20.588 & 0.818 & 0.143 & 98.949 & 0.129 \\
KinD++~\cite{zhang2021beyond} & 17.752 & 0.758 & 0.198 & 80.625 & 0.124 & 17.660 & 0.761 & 0.218 & 99.504 & 0.136 \\
RetinexNet~\cite{wei2018deep} & 17.558 & 0.651 & 0.379 & 150.500 & 0.117 & 16.097 & 0.401 & 0.543 & 158.988 & 0.131 \\
DRBN~\cite{yang2020fidelity} & 16.777 & 0.730 & 0.345 & 125.238 & 0.126 & 18.466 & 0.768 & 0.352 & 110.984 & 0.138 \\
DLN~\cite{wang2020lightening} & 21.946 & 0.807 & 0.148 & 85.729 & 0.083 & 17.878 & 0.693 & 0.300 & 95.625 & 0.121 \\
URetinex-Net~\cite{wu2022uretinex} & 21.328 & 0.835 & - & 59.000 & 0.083 & - & - & - & 48.360 & - \\
EnlightenGAN~\cite{jiang2021enlightengan} & 17.606 & 0.653 & 0.372 & 105.590 & 0.135 & 18.676 & 0.678 & 0.364 & 92.562 & 0.108 \\
GDP~\cite{fei2023generative} & 15.821 & 0.541 & 0.338 & 120.478 & 0.147 & 14.412 & 0.497 & 0.363 & 100.215 & 0.142 \\
Restormer~\cite{zamir2022restormer} & 22.156 & 0.817 & 0.151 & 72.153 & 0.078 & 21.236 & 0.820 & 0.191 & 70.714 & 0.085 \\
LLFormer~\cite{wang2023ultra} & 23.649 & 0.819 & 0.169 & 76.960 & 0.063 & 20.154 & 0.809 & 0.207 & 70.165 & 0.105 \\
Retinexformer~\cite{cai2023retinexformer} & 23.386 & 0.833 & 0.140 & 70.287 & 0.067 & 22.254 & 0.831 & 0.184 & 65.031 & 0.074 \\
NeRCo~\cite{yang2023implicit} & 22.946 & 0.786 & 0.146 & 80.742 & 0.069 & 18.490 & 0.633 & 0.414 & 200.438 & 0.111 \\
CUE~\cite{zheng2023empowering} & 21.670 & 0.775 & 0.224 & 93.861 & 0.079 & 18.053 & 0.753 & 0.347 & 108.284 & 0.129 \\
C-Retinex~\cite{xu2024cretinex} & 19.866 & 0.807 & 0.196 & 90.185 & 0.096 & 18.265 & 0.789 & 0.285 & 106.526 & 0.122 \\
HVI-CIDNet~\cite{yan2025hvi} & 23.986 & 0.854 & 0.092 & 72.448 & 0.065 & 23.629 & 0.868 & 0.107 & 63.727 & 0.061 \\
IR-CM~\cite{gong2024ir} & 21.878 & 0.821 & 0.139 & 106.643 & 0.085 & 22.842 & 0.841 & 0.145 & 97.367 & 0.071 \\
CLE Diffusion~\cite{yin2023cle} & 21.282 & 0.794 & 0.157 & 74.540 & 0.080 & 21.995 & 0.798 & 0.191 & 79.859 & 0.074 \\
QuadPrior~\cite{wang2024zero} & 18.771 & 0.781 & 0.209 & 80.185 & 0.112 & 20.471 & 0.811 & 0.198 & 69.945 & 0.085 \\
PyDiff~\cite{zhou2023pyramid} & 23.275 & 0.858 & 0.107 & 50.190 & 0.074 & 21.940 & 0.841 & 0.178 & 81.571 & 0.072 \\
AnlightenDiff~\cite{chan2024anlightendiff} & 23.500 & 0.840 & 0.120 & 53.200 & 0.075 & 22.100 & 0.850 & 0.115 & 48.900 & 0.070 \\
GSAD~\cite{hou2023global} & 22.978 & 0.851 & 0.103 & 49.881 & 0.076 & 20.190 & 0.847 & 0.112 & 46.765 & 0.102 \\
Diff-Retinex~\cite{yi2023diff} & 21.980 & 0.863 & 0.098 & \textbf{47.850} & 0.076 & 21.873 & 0.864 & 0.131 & 47.750 & 0.080 \\
Diff-Retinex++~\cite{yi2025diff} & 24.667 & 0.867 & 0.101 & 50.771 & 0.062 & 23.413 & 0.872 & 0.134 & 49.592 & 0.064 \\
ReDDiT~\cite{lan2025efficient} (2 steps) & 23.654 & 0.847 & 0.109 & 79.540 & 0.071 & 23.576 & 0.858 & 0.114 & 69.093 & 0.057 \\
\textbf{Consist-Retinex (Ours, one-step)} & 22.450$\pm$0.31 & 0.826$\pm$0.004 & 0.126$\pm$0.002 & 62.586$\pm$0.74 & 0.078$\pm$0.001 & 25.512$\pm$0.18 & 0.880$\pm$0.003 & 0.106$\pm$0.002 & 44.726$\pm$0.58 & 0.0493$\pm$0.0008 \\
\textbf{Consist-Retinex++ (Ours, one-step)} & \textbf{24.962$\pm$0.12} & \textbf{0.875$\pm$0.002} & \textbf{0.089$\pm$0.001} & 52.793$\pm$0.66 & \textbf{0.057$\pm$0.001} & \textbf{26.031$\pm$0.09} & \textbf{0.898$\pm$0.002} & \textbf{0.093$\pm$0.001} & \textbf{41.329$\pm$0.45} & \textbf{0.0452$\pm$0.0006} \\
\hline
\end{tabular}}

\end{table*}

\section{Theoretical Analysis}
\label{sec_m:theory}

We provide endpoint-oriented theoretical insights for Consist-Retinex. The corresponding proofs are provided in Section~S5 of the supplementary material.

\subsection{Endpoint Error for One-Step Inference}

\begin{theorem}[Endpoint fixed-point error bound]
\label{main_thm:convergence}
For each Retinex component $Z\in\{R,L\}$, let
$X_{\text{tr}}^Z=Z_n+\sigma_{\max}\xi_Z$ denote the high-noise endpoint used by
the fixed-point loss and $X_{\text{inf}}^Z=\sigma_{\max}\xi_Z$ denote the
pure-noise endpoint used by one-step inference, with
$\xi_Z\sim\mathcal{N}(0,I)$, and let $C_l^Z$ be the corresponding low-light TDN
condition ($C_l^R=R_l$, $C_l^L=L_l$). Define
\begin{equation}
\begin{aligned}
\epsilon_{\text{fixed}}
={}&
\lambda_{\text{fixed}}\,p_{\text{large}} \\
&\times\sum_{Z\in\{R,L\}}
\mathbb{E}\Big[
\big\|f_{\theta}^{Z}(X_{\text{tr}}^Z,\sigma_{\max}\mid C_l^Z)-Z_n\big\|_2^2
\Big],
\end{aligned}
\end{equation}
and the inference-prior mismatch
\begin{equation}
\begin{aligned}
\Gamma_{\text{prior}}
={}&
\sum_{Z\in\{R,L\}}
\mathbb{E}\Big[
\big\|f_{\theta}^{Z}(X_{\text{inf}}^Z,\sigma_{\max}\mid C_l^Z) \\
&\qquad\quad
-f_{\theta}^{Z}(X_{\text{tr}}^Z,\sigma_{\max}\mid C_l^Z)\big\|_2
\Big].
\end{aligned}
\end{equation}
and the TDN reconstruction residual
\begin{equation}
\epsilon_{\text{TDN}}
:=
\mathbb{E}\bigl[\|R_n\odot L_n-I_n\|_F\bigr].
\end{equation}
Then the one-step output $\hat I_n=\hat R_n\odot\hat L_n$ satisfies
\begin{equation}
W_1(\hat I_n,I_n)
\leq
\sqrt{\frac{2\epsilon_{\text{fixed}}}
{\lambda_{\text{fixed}}p_{\text{large}}}}
+\Gamma_{\text{prior}}
+\epsilon_{\text{TDN}}.
\label{eq_main:main_bound}
\end{equation}
The last term is the reconstruction residual of the frozen first-stage
decomposition; it vanishes when the TDN factorization is exactly reconstructive
($I_n=R_n\odot L_n$) and is unaffected by consistency training. Among the three
terms, the one the consistency stage directly controls is therefore the
high-noise fixed-point error; temporal consistency acts as trajectory
regularization rather than an additional ODE-discretization bottleneck for
evaluating the endpoint.
\end{theorem}

The mismatch term arises because training perturbs $Z_n$ while inference starts
from pure noise. Under EDM preconditioning, the next lemma shows that this term
is suppressed at the endpoint by an explicit $O(10^{-3})$ factor, leaving the
high-noise fixed-point error as the dominant quantity in
Theorem~\ref{main_thm:convergence}.

\begin{lemma}[Explicit EDM contraction of the prior-mismatch term]
\label{main_lem:edm_contraction}
Assume that for $\sigma=\sigma_{\max}$ and any low-light condition $C_l^Z$,
the EDM inner network $F_\theta(\,\cdot\,,\sigma_{\max}\mid C_l^Z)$ is
$L_F$-Lipschitz in its first argument on the segment joining
$c_{\text{in}}(\sigma_{\max})X_{\text{tr}}^Z$ and
$c_{\text{in}}(\sigma_{\max})X_{\text{inf}}^Z$. Then
\begin{equation}
\Gamma_{\text{prior}}
\;\leq\;
\begin{aligned}
&\kappa(\sigma_{\max};L_F)
\sum_{Z\in\{R,L\}}\mathbb{E}\|Z_n\|_2, \\
&\text{where}\quad
\kappa(\sigma;L_F)
:=
\frac{\sigma_{\text{data}}\bigl(\sigma_{\text{data}}+L_F\,\sigma\bigr)}
     {\sigma^{2}+\sigma_{\text{data}}^{2}}.
\end{aligned}
\label{eq_main:edm_contraction}
\end{equation}
For our setting $\sigma_{\max}=80$ and $\sigma_{\text{data}}=0.5$,
\begin{equation}
\kappa(\sigma_{\max};L_F)
\;\approx\;
3.9\times 10^{-5}\;+\;6.25\times 10^{-3}\,L_F,
\end{equation}
so for networks whose effective local Lipschitz constant near the endpoint
satisfies $L_F=O(1)$, the prior-mismatch contribution to
Eq.~\ref{eq_main:main_bound} is bounded by an EDM endpoint factor on the order
of $10^{-3}\cdot\mathbb{E}\|Z_n\|_2$. This is a conditional statement: $L_F$ is
assumed rather than estimated for the trained network. Under that condition the
remaining quantity controlling one-step inference accuracy is the high-noise
fixed-point error $\epsilon_{\text{fixed}}$, exactly the term that the proposed
dual objective and noise-emphasized sampling are designed to drive down.
\end{lemma}

\subsection{Necessity of the Fixed-Point Term}
\label{subsec:necessity}

The temporal-consistency objective, on its own, has a large null space: any
predictor that depends only on the conditioning input is trivially
self-consistent. This formalizes why dual-objective training is required for
conditional Retinex enhancement.

\begin{proposition}[Degeneracy of consistency-only training]
\label{main_prop:degeneracy}
For each Retinex component $Z\in\{R,L\}$, let
$\mathcal{F}_{\text{const}}^{Z}
=\bigl\{\tilde f_{g}(x,\sigma\mid C_l^Z)\;{:=}\;g(C_l^Z)
\,\big|\,g:\mathcal{C}\!\to\!\mathcal{X}\bigr\}$
denote the family of predictors that depend only on the low-light condition.
Then for every $\tilde f_g\in\mathcal{F}_{\text{const}}^Z$,
\begin{equation}
\mathcal{L}_{\text{consist}}(\tilde f_g)
\;=\;0
\qquad\text{exactly, regardless of $g$.}
\label{eq_main:degenerate_consist}
\end{equation}
In particular, the temporal consistency objective alone admits an
infinite-dimensional family of zero-loss minimizers, only one of which yields
the correct one-step Retinex output. Among $\mathcal{F}_{\text{const}}^Z$,
this minimizer is uniquely selected by $\mathcal{L}_{\text{fixed}}$: the
constant-in-$(x,\sigma)$ predictor minimizing the fixed-point loss is
$g^{\star}(C_l^Z)=\mathbb{E}[Z_n\mid C_l^Z]$.
\end{proposition}

Thus, self-consistency alone cannot select the desired enhanced component from
condition-only zero-loss maps. The fixed-point term provides this selection,
explaining the collapse observed when it is removed in Table~\ref{tab:ablation}
(9.34~dB on LOL).

\subsection{Propagation of Endpoint Anchoring}

\begin{proposition}[Consistency propagates high-noise anchoring]
\label{main_prop:anchoring}
Fix one Retinex component $Z\in\{R,L\}$ and one conditional trajectory
$\{X_t^Z\}_{t=1}^T$ associated with the same low-light TDN condition $C_l^Z$, where
$X_T^Z = Z_n + \sigma_{\max}\xi_Z$ and $\xi_Z\sim\mathcal{N}(0,I)$. Define the
adjacent-time consistency errors
\begin{equation}
\begin{aligned}
\eta_t^Z
={}&
\mathbb{E}\Big[
\big\|f_t^Z(X_t^Z\mid C_l^Z)-f_{t+1}^Z(X_{t+1}^Z\mid C_l^Z)\big\|_2^2
\Big], \\
& t=1,\dots,T-1,
\end{aligned}
\end{equation}
and the terminal anchoring error
\begin{equation}
a_T^Z
=
\mathbb{E}\left[
\left\|f_T^Z(X_T^Z\mid C_l^Z)-Z_n\right\|_2^2
\right].
\end{equation}
Then for every $t\in\{1,\dots,T\}$,
\begin{equation}
\mathbb{E}\left[
\left\|f_t^Z(X_t^Z\mid C_l^Z)-Z_n\right\|_2
\right]
\le
\sqrt{a_T^Z}
\;+\;
\sum_{s=t}^{T-1}\sqrt{\eta_s^Z}
\label{eq_m:anchor_path_1}
\end{equation}
and therefore
\begin{equation}
\begin{aligned}
\mathbb{E}\Big[
\big\|f_t^Z(X_t^Z\mid C_l^Z)-Z_n\big\|_2
\Big]
\le{}&
\sqrt{a_T^Z} \\
&+
\sqrt{T-t}\Big(\sum_{s=t}^{T-1}\eta_s^Z\Big)^{1/2}.
\end{aligned}
\label{eq_m:anchor_path_2}
\end{equation}
In particular, exact self-consistency ($\eta_t^Z=0$ for all $t$) plus exact terminal anchoring ($a_T^Z=0$) implies
$f_t^Z(X_t^Z\mid C_l^Z)=Z_n$ almost surely for every point on the trajectory.
\end{proposition}

Proposition~\ref{main_prop:anchoring} clarifies the role of the dual objective from a consistency-model viewpoint: temporal consistency keeps predictions aligned \emph{with each other}, while the high-noise fixed-point term selects \emph{which} trajectory-constant value they should share. Without anchoring, a consistency model may be self-consistent yet converge to an incorrect conditional target.

\subsection{High-Noise Sample Allocation}

We next quantify the sampling advantage of the noise-emphasized fixed-point
objective without claiming a full neural-network generalization bound.
\begin{theorem}[High-noise sample allocation]
\label{main_thm:sample_complexity}
Let $N_{\text{large}}$ be the number of fixed-point training samples whose noise
level lies in $[0.95\sigma_{\max},\sigma_{\max}]$ after $N_{\text{train}}$
draws. If each draw lands in this interval with probability $p_{\text{large}}$,
then $\mathbb{E}[N_{\text{large}}]=p_{\text{large}}N_{\text{train}}$ and, for
any target count $m$ and failure probability $\delta$, it suffices to use
\begin{equation}
N_{\text{train}}
\geq
\frac{2m}{p_{\text{large}}}
+
\frac{8}{p_{\text{large}}}\log\frac{1}{\delta}
\label{eq_m:complexity}
\end{equation}
to ensure $N_{\text{large}}\geq m$ with probability at least $1-\delta$.
Equivalently, the required total samples scale as $p_{\text{large}}^{-1}$ for a
fixed required number of high-noise examples. In our setting,
$p_{\text{large}}\approx0.950$ under Eq.~\ref{eq:bimodal_sampling}, compared
with $p_{\text{large}}^{\text{std}}\approx0.0048$ for full-range log-uniform
sampling.
\end{theorem}

\subsection{Concentration of the High-Noise Empirical Risk}
\label{main_subsec:concentration}

The sample-allocation theorem counts how many high-noise samples appear; we
now translate this directly into a concentration radius for the empirical
endpoint risk under either sampling scheme. This sharpens the qualitative
claim that noise-emphasized sampling improves \emph{statistical efficiency},
not just the expected count.

\begin{theorem}[Concentration of the high-noise empirical risk]
\label{main_thm:concentration}
Fix a predictor $\theta$ and let
$\ell(\theta;\sigma,\xi,C):=\bigl\|f_\theta(x_0+\sigma\xi,\sigma\mid C)-x_0\bigr\|_2^2$
denote the per-sample fixed-point loss for any clean target $x_0$, condition
$C$, noise scale $\sigma$, and Gaussian noise $\xi\sim\mathcal{N}(0,I)$.
Assume that $\ell$ is uniformly bounded by some $B>0$ on the training
domain. Let $\pi$ be a noise-sampling distribution with
$p:=\pi(I_{\text{large}})>0$ on the high-noise interval
$I_{\text{large}}=[0.95\sigma_{\max},\sigma_{\max}]$, and define the
high-noise population and empirical risks
\begin{equation}
\begin{aligned}
R_{\text{large}}(\theta)
&:=
\mathbb{E}_{(\sigma,\xi,x_0,C)}\bigl[\ell(\theta;\sigma,\xi,C)
\,\big|\,\sigma\in I_{\text{large}}\bigr], \\
\hat R_{\text{large}}^{\pi}(\theta;N)
&:=
\frac{1}{N_{\text{large}}}
\sum_{i:\sigma_i\in I_{\text{large}}}
\ell(\theta;\sigma_i,\xi_i,C_i),
\end{aligned}
\end{equation}
where $\{(\sigma_i,\xi_i,x_{0,i},C_i)\}_{i=1}^{N}$ are $N$ i.i.d.\ training
draws under $\pi$ and $N_{\text{large}}$ is the number of these whose
$\sigma_i\in I_{\text{large}}$. Then for any $\delta\in(0,1)$, with
probability at least $1-\delta$ jointly over the sampling and the loss
realization,
\begin{equation}
\bigl|\hat R_{\text{large}}^{\pi}(\theta;N)-R_{\text{large}}(\theta)\bigr|
\;\leq\;
B\sqrt{\frac{\log(4/\delta)}{N\,p}},
\label{eq_m:concentration_bound}
\end{equation}
provided $N\geq 8\,p^{-1}\log(2/\delta)$.
\end{theorem}

\subsection{Implications for One-Step Inference}

Theorem~\ref{main_thm:convergence}, Lemma~\ref{main_lem:edm_contraction},
Proposition~\ref{main_prop:degeneracy}, Proposition~\ref{main_prop:anchoring},
Theorem~\ref{main_thm:sample_complexity}, and
Theorem~\ref{main_thm:concentration} jointly clarify the role of the proposed
training design:

\noindent\textbf{Endpoint accuracy is the dominant term.} Theorem~\ref{main_thm:convergence} bounds the one-step Wasserstein-1 error by a fixed-point error at $\sigma_{\max}$, the prior-mismatch term $\Gamma_{\text{prior}}$, and the reconstruction residual $\epsilon_{\text{TDN}}$ of the frozen decomposition stage. Lemma~\ref{main_lem:edm_contraction} converts the latter into an explicit EDM-preconditioning factor $\kappa(\sigma_{\max};L_F)\approx 6.25\times10^{-3}\,L_F+3.9\times10^{-5}$. Hence the prior-mismatch is multiplicatively suppressed by an order-$10^{-3}$ constant, and the only quantity that remains free to optimize is the high-noise fixed-point error.

\noindent\textbf{Necessity of the fixed-point term, formalized.} Proposition~\ref{main_prop:degeneracy} shows that any predictor depending only on the conditioning input lies in the null space of $\mathcal{L}_{\text{consist}}$. This formalizes why removing $\mathcal{L}_{\text{fixed}}$ cannot just yield a worse model -- it removes the only mechanism that selects the correct trajectory-constant value out of an infinite-dimensional zero-loss family, matching the 9.34~dB collapse in Table~\ref{tab:ablation}. Standard consistency training~\cite{song2023consistency} lacks this term because unconditional generation does not use paired ground-truth targets.

\noindent\textbf{Propagation of conditional supervision.} Proposition~\ref{main_prop:anchoring} shows that once the prediction at the high-noise endpoint is anchored to the correct target, temporal self-consistency propagates that supervision backward along the whole conditional trajectory. This is specific to consistency models: the temporal objective alone can enforce trajectory-wise agreement, but only the endpoint anchor (selected by $\mathcal{L}_{\text{fixed}}$ via Proposition~\ref{main_prop:degeneracy}) determines whether the shared value is the desired enhanced component or an incorrect fixed point.

\noindent\textbf{Benefit of noise-emphasized sampling.} The $p_{\text{large}}^{-1}$ dependence in~\eqref{eq_m:complexity} shows why high-noise samples are statistically valuable for one-step inference. Theorem~\ref{main_thm:concentration} sharpens this from a sample-count statement into an explicit $1/\sqrt{Np}$ concentration radius, so that the proposed sampler tightens the empirical-risk estimate at the endpoint by a factor of $\sqrt{p_{\text{large}}^{\text{NE}}/p_{\text{large}}^{\text{std}}}\approx14$ for the same total budget. Our ablation confirms this: reducing the high-noise probability substantially lowers PSNR on both LOL and VE-LOL-L.

\noindent\textbf{Scope of the analysis.} The endpoint bound, contraction lemma, allocation result, and concentration inequality together explain why high-noise anchoring with paired ground-truth supervision is necessary for one-step inference, but they do not constitute a full generalization theorem for the learned neural network. Such a result would require explicit capacity control for the model class and optimizer-dependent assumptions.

\section{Experiments}
\label{sec:experiments}

\subsection{Implementation Details and Datasets}
\label{subsec:implementation}

\noindent\textbf{Implementation Details.}
We implement Consist-Retinex in PyTorch~\cite{paszke2019pytorch} and train on 3 NVIDIA RTX A6000 GPUs. Both consistency models are trained for 200K iterations with batch size 8 using AdamW~\cite{loshchilov2017decoupled} on $160\times160$ patches. We use $\sigma_{\min}=0.002$, $\sigma_{\max}=80$, $\tau=0.95$, $p_{\text{large}}=0.95$, $\lambda_{\text{consist}}=1.0$, and $\lambda_{\text{fixed}}=0.3$. The consistency-stage hyperparameters and the full training protocol are given in Section~S1 of the supplementary material (Table~S1).

\textbf{Metric protocol.} All full-reference metrics in Table~\ref{tab:lol_velol_results} are computed on clipped RGB outputs mapped to the standard 8-bit $[0,255]$ range; we do not use GT-mean adjustment or 6-bit quantization. LPIPS uses the AlexNet backbone from the standard LPIPS implementation. For our rows, the reported values are mean$\pm$standard deviation over six stochastic one-step samplings with the same trained checkpoint.
\noindent\textbf{Datasets.}
\label{app:data_baselines}

\noindent\textbf{Datasets.}
For paired evaluation, we use the LOL dataset~\cite{wei2018deep}, which contains 485 training images and 15 test images at $600 \times 400$ resolution, and VE-LOL-L~\cite{liu2021benchmarking}, comprising 689 training images and 100 test images with diverse real-world scenes. To probe cross-dataset behavior beyond paired supervision, we further test on two unpaired datasets: VV, consisting of 24 low-light images for zero-shot evaluation, and DICM, which includes 64 images for cross-dataset testing.

\noindent\textbf{Compared methods.}
We compare Consist-Retinex against representative methods from multiple categories: traditional approaches (LIME~\cite{guo2016lime}, JED~\cite{ren2018joint}), Retinex-based methods (RetinexNet~\cite{wei2018deep}, KinD~\cite{zhang2019kindling}, KinD++~\cite{zhang2021beyond}, URetinex~\cite{wu2022uretinex}, C-Retinex~\cite{xu2024cretinex}), Transformer-based methods (Restormer~\cite{zamir2022restormer}, LLFormer~\cite{wang2023ultra}, Retinexformer~\cite{cai2023retinexformer}), diffusion-based approaches (Diff-Retinex (1000 steps)~\cite{yi2023diff}, Diff-Retinex++ (1000 steps)~\cite{yi2025diff}, ReDDiT (2 steps)~\cite{lan2025efficient}, CLE Diffusion~\cite{yin2023cle}, PyDiff~\cite{zhou2023pyramid}, GSAD~\cite{hou2023global}), consistency-based restoration (IR-CM~\cite{gong2024ir}), and other enhancement techniques (Zero-DCE~\cite{guo2020zero}, EnlightenGAN~\cite{jiang2021enlightengan}, HVI~\cite{yan2025hvi}).

\subsection{Comparison with State-of-the-Art Methods}
\label{subsec:comparison}

\subsubsection{Quantitative Results on LOL and VE-LOL-L}
\label{subsubsec:quantitative_reference}
\noindent\textbf{Full-reference metrics.}
We report PSNR, SSIM~\cite{wang2004image}, LPIPS~\cite{zhang2018unreasonable}, FID~\cite{heusel2017gans}, and MAE in Table~\ref{tab:lol_velol_results}. The default one-step Consist-Retinex is the clean comparison for the proposed training strategy under the base Retinex backbone; it is not claimed to be SOTA on LOL, but it improves step-efficient Retinex enhancement and gives strong VE-LOL-L results with one sampling step. Consist-Retinex++ uses the same training strategy in the Diff-Retinex++-style backbone/augmentation setting. This controlled same-architecture comparison is important: relative to Diff-Retinex++~\cite{yi2025diff}, the ++ variant changes the training/sampling algorithm while keeping the stronger architecture family, and improves PSNR/SSIM/LPIPS/MAE on both paired benchmarks. We therefore separate the default row, which isolates the training-strategy contribution more directly, from the ++ row, which shows that the proposed algorithm remains beneficial when the architecture and augmentation are matched to a stronger baseline. The only metric where our method does not rank first is LOL FID, where Diff-Retinex~\cite{yi2023diff} remains lower under the reported metric implementation.

\noindent\textbf{LOL vs.\ VE-LOL-L performance.}
The higher PSNR on VE-LOL-L (25.51 dB) compared to LOL (22.45 dB) stems from inherent dataset characteristics rather than model bias: (1) \textit{Higher image quality}---VE-LOL-L consists of real-captured images with better overall quality and less severe degradation than LOL's mixture of synthetic and heavily degraded samples~\cite{liu2021benchmarking}; (2) \textit{Larger test set}---VE-LOL-L's 100 test images vs.\ LOL's 15 provide more statistically reliable metrics, reducing variance from outlier cases.

\subsubsection{Backbone and Augmentation Variant: Consist-Retinex++}
\label{subsubsec:plus_variant}

The Consist-Retinex++ row of Table~\ref{tab:lol_velol_results} is included to remove a common ambiguity in comparing methods with different architectures. The default Consist-Retinex row is the direct evidence for the proposed algorithm under the base Retinex backbone. Consist-Retinex++ instead keeps the Diff-Retinex++-style restoration backbone and augmentation recipe, including the Retinex-aware DRNet/RMoE design, high-sequence concatenation, brightness augmentation, and low-light mixup, but replaces the iterative diffusion objective with our dual-objective consistency training and noise-emphasized one-step sampling. Therefore, the comparison between Diff-Retinex++ and Consist-Retinex++ is the matched-architecture comparison: under the same stronger architecture family, our training/sampling algorithm gives stable gains on both LOL and VE-LOL-L while reducing inference to one step. We do not use this row to claim that the full PSNR gain is caused by consistency training alone; rather, it shows that the proposed algorithm remains beneficial when architecture and augmentation effects are controlled.

\subsubsection{No-Reference Evaluation on VV and DICM}
\label{subsubsec:no_reference}

For unpaired DICM and VV, we report PI~\cite{blau2018perception} and NIQE~\cite{mittal2012making} in Table~\ref{tab:dicm_vv_results} (lower is better). Consist-Retinex obtains the best PI on DICM and remains competitive on VV, although DiffLL and LightenDiffusion still achieve lower no-reference scores on VV. We therefore treat these unpaired results as diagnostic evidence of cross-dataset behavior rather than as a broad generalization claim.
\begin{table*}[t]
\centering
\caption{No-reference evaluation on DICM and VV datasets. Lower is better.}
\label{tab:dicm_vv_results}

\begin{minipage}{0.48\textwidth}
\centering
\caption*{(a) DICM dataset}
\begin{tabular}{l|cc}
\hline
Method & PI$\downarrow$ & NIQE$\downarrow$ \\
\hline
CLE Diffusion~\cite{yin2023cle} & 3.361 & 4.505 \\
DiffLL~\cite{jiang2023low} & 2.936 & 3.636 \\
GDP~\cite{fei2023generative} & 3.552 & 4.358 \\
Diff-Retinex~\cite{yi2023diff} & 3.394 & 4.361 \\
Diff-Retinex++~\cite{yi2025diff} & 3.203 & \textbf{3.514} \\
PyDiffusion~\cite{zhou2023pyramid} & 3.792 & 4.499 \\
LightenDiffusion~\cite{jiang2024lightendiffusion} & 3.144 & 3.724 \\
\textbf{Consist-Retinex (Ours)} & \textbf{2.932} & 3.826 \\
\hline
\end{tabular}
\end{minipage}
\hfill
\begin{minipage}{0.48\textwidth}
\centering
\caption*{(b) VV dataset}
\begin{tabular}{l|cc}
\hline
Method & PI$\downarrow$ & NIQE$\downarrow$ \\
\hline
CLE Diffusion~\cite{yin2023cle} & 3.470 & 3.240 \\
DiffLL~\cite{jiang2023low} & \textbf{2.351} & \textbf{2.869} \\
GDP~\cite{fei2023generative} & 3.431 & 4.683 \\
Diff-Retinex~\cite{yi2023diff} & 3.350 & 3.087 \\
Diff-Retinex++~\cite{yi2025diff} & 3.643 & 3.819 \\
PyDiffusion~\cite{zhou2023pyramid} & 3.678 & 4.360 \\
LightenDiffusion~\cite{jiang2024lightendiffusion} & 2.558 & 2.941 \\
\textbf{Consist-Retinex (Ours)} & 2.663 & 2.985 \\
\hline
\end{tabular}
\end{minipage}

\end{table*}
\subsubsection{Qualitative Comparison}
\label{subsubsec:qualitative}

\begin{figure*}[!t]
    \centering
    \includegraphics[width=\linewidth]{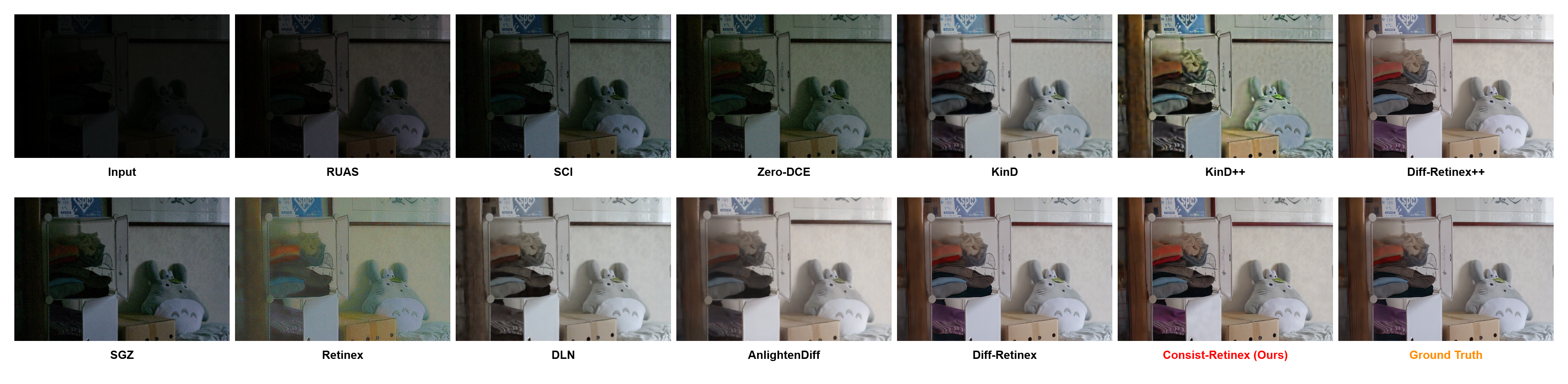}
    \caption{Qualitative comparison with representative low-light image enhancement methods on the LOL dataset.}
    \label{fig1:consistency_mapping}
\end{figure*}

\begin{figure*}[!t]
    \centering
    \includegraphics[width=\linewidth]{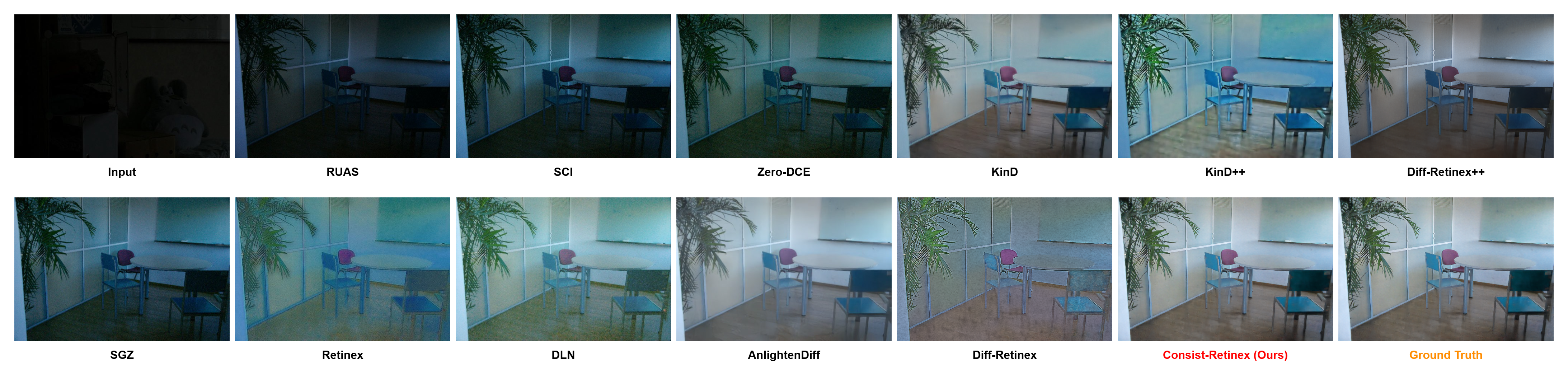}
    \caption{Qualitative comparison with representative low-light image enhancement methods on the VE-LOL-L dataset.}
    \label{fig2:consistency_mapping}
\end{figure*}

Figs.~\ref{fig1:consistency_mapping} and~\ref{fig2:consistency_mapping} show representative qualitative results on LOL and VE-LOL-L, DICM/VV examples (Fig.~S2) and component-level RDA/IDA visual diagnostics (Figs.~S3 and~S4) are given in the supplementary material. Overall, the paired examples show faithful illumination and clear local structures, while unpaired results remain case-dependent, consistent with Table~\ref{tab:dicm_vv_results}.

\subsection{Comparison with Accelerated Diffusion Sampling}

To further validate the step-efficiency advantage of our consistency-based approach, we compare Consist-Retinex with Diff-Retinex using DDIM~\cite{song2021denoising} accelerated sampling. DDIM is a deterministic sampling method that can reduce the number of inference steps while maintaining reasonable quality.

\textbf{Experimental Setup.} We evaluate Diff-Retinex with DDIM sampling at different step configurations on both LOL and VE-LOL-L datasets, comparing against our one-step Consist-Retinex. All models use the same pre-trained Diff-Retinex checkpoints reported in the main paper. This comparison focuses on sampling steps and reconstruction quality; it is not intended as a controlled wall-clock benchmark because exact runtime depends on hardware, batch size, codebase, and implementation optimizations.

\textbf{Results.} Table~\ref{tab:ddim_comparison} presents the quantitative comparison on both datasets. While DDIM sampling reduces inference steps from 1000 to 5, it incurs substantial performance degradation on LOL: PSNR drops from 21.98~dB to 19.82~dB (2.16~dB loss), and FID increases from 47.85 to 68.04 (20.19-point degradation). Similar trends are observed on VE-LOL-L, where 5-step DDIM sampling achieves 20.45~dB PSNR and 72.06 FID, significantly worse than the 1000-step baseline (21.87~dB, 47.75 FID). This demonstrates the inherent trade-off in ODE solver-based acceleration methods.

In contrast, Consist-Retinex achieves 22.45~dB PSNR and 62.59 FID on LOL with single-step inference, outperforming DDIM 5-step sampling by 2.63~dB in PSNR and 5.45 points in FID while requiring 5$\times$ fewer sampling steps. On VE-LOL-L, our method achieves 25.51~dB PSNR and 44.73 FID, surpassing DDIM 5-step by 5.06~dB and 27.33 FID points. Compared to the full 1000-step Diff-Retinex, our method reduces the number of sampling steps by 1000$\times$ while maintaining strong performance, validating that consistency training provides an effective path to step-efficient high-quality enhancement.

\begin{table*}[t]
\centering
\caption{Comparison with DDIM accelerated sampling on LOL and VE-LOL-L datasets. The table reports sampling-step efficiency and quality; wall-clock runtime is not directly comparable without a controlled implementation-level benchmark.}
\label{tab:ddim_comparison}
\resizebox{\textwidth}{!}{
\begin{tabular}{l|c|ccc|ccc}
\hline
\multirow{2}{*}{Method} & \multirow{2}{*}{Steps} & \multicolumn{3}{c|}{LOL Dataset} & \multicolumn{3}{c}{VE-LOL-L Dataset} \\
\cline{3-8}
& & PSNR$\uparrow$ & SSIM$\uparrow$ & FID$\downarrow$ & PSNR$\uparrow$ & SSIM$\uparrow$ & FID$\downarrow$ \\
\hline
Diff-Retinex (DDPM) & 1000 & 21.98 & 0.863 & 47.85 & 21.873 & 0.864 & 47.75 \\
Diff-Retinex (DDIM) & 50 & 20.95 & 0.841 & 58.32 & 21.38 & 0.805 & 59.68 \\
Diff-Retinex (DDIM) & 10 & 20.23 & 0.815 & 63.78 & 20.89 & 0.785 & 65.42 \\
Diff-Retinex (DDIM) & 5 & 19.82 & 0.798 & 68.04 & 20.45 & 0.768 & 72.06 \\
\hline
\textbf{Consist-Retinex (Ours)} & \textbf{1} & \textbf{22.45} & \textbf{0.826} & \textbf{62.59} & \textbf{25.51} & \textbf{0.880} & \textbf{44.73} \\
\hline
\end{tabular}
}
\end{table*}

\textbf{Analysis.} The superior performance of Consist-Retinex over DDIM sampling stems from fundamental differences in their acceleration mechanisms:

\noindent\textbf{(1) Training vs. Inference Acceleration:} DDIM accelerates only inference by skipping intermediate denoising steps, but the model remains trained for 1000-step trajectories. This mismatch causes quality degradation. Consist-Retinex explicitly trains for one-step generation, aligning training and inference dynamics.

\noindent\textbf{(2) Conditional Mapping Design:} Our noise-emphasized sampling and dual-objective loss specifically optimize the $\sigma = \sigma_{\max} \rightarrow \sigma = 0$ mapping critical for one-step conditional enhancement. DDIM lacks such task-specific training adaptations.

\noindent\textbf{(3) Efficiency Gains:} Beyond fewer inference steps, Consist-Retinex uses only 1/8 as many effective training samples (1.6M vs. 12.8M), demonstrating a clear training-cost advantage in the reported setup. We report wall-clock latency only for our implementation and avoid converting step counts directly into wall-clock speedups.

\noindent\textbf{(4) Dataset-Dependent Performance:} The performance gap between our method and DDIM sampling is more pronounced on VE-LOL-L (5.06~dB improvement) than on LOL (2.39~dB improvement), suggesting that consistency training is particularly beneficial for complex, diverse real-world scenarios where one-step conditional mapping is more challenging.

These results suggest that consistency modeling provides an effective route for step-efficient conditional generation, avoiding the quality degradation observed with post-hoc ODE solver acceleration of standard diffusion models in our experiments.

\subsection{Ablation Study}
\label{subsec:ablation}

\begin{table}[t]
\centering
\caption{Ablation study on LOL and VE-LOL-L datasets. Consist: temporal consistency loss; GT: ground-truth alignment loss; NE: noise-emphasized sampling.}
\label{tab:ablation}
\resizebox{\columnwidth}{!}{
\begin{tabular}{ccc|ccc|ccc}
\hline
\multirow{2}{*}{Consist} & \multirow{2}{*}{GT} & \multirow{2}{*}{NE} & \multicolumn{3}{c|}{LOL Dataset} & \multicolumn{3}{c}{VE-LOL-L Dataset} \\
\cline{4-9}
 &  &  & PSNR$\uparrow$ & SSIM$\uparrow$ & LPIPS$\downarrow$ & PSNR$\uparrow$ & SSIM$\uparrow$ & LPIPS$\downarrow$ \\ 
\hline
\xmark & \cmark & \cmark & 17.14 & 0.685 & 0.246 & 18.26 & 0.764 & 0.224 \\
\cmark & \xmark & \xmark & 9.34 & 0.485 & 0.385 & 10.51 & 0.528 & 0.362 \\
\cmark & \cmark & \xmark & 18.82 & 0.752 & 0.189 & 20.05 & 0.821 & 0.165 \\
\cmark & \cmark & \cmark & \textbf{22.45} & \textbf{0.826} & \textbf{0.126} & \textbf{25.51} & \textbf{0.880} & \textbf{0.106} \\
\hline
\end{tabular}}
\end{table}

\subsubsection{Component Ablation}
We first ablate the ground-truth alignment loss and noise-emphasized sampling on LOL and VE-LOL-L under the same training setting. Table~\ref{tab:ablation} shows that temporal consistency alone collapses to 9.34/10.51 dB PSNR on LOL/VE-LOL-L, while GT alignment with noise-emphasized sampling but without consistency remains limited at 17.14/18.26 dB. Combining consistency with fixed-point alignment improves PSNR to 18.82/20.05 dB, and adding noise-emphasized sampling further raises it to 22.45/25.51 dB with the best LPIPS. These results support both the anchoring role of the fixed-point term and the benefit of endpoint-focused sampling.
\begin{table}[t]
\centering
\caption{Matched cross-setting ablation on LOL. Each row is one algorithmic
configuration; the two result columns apply that configuration to the base
Retinex backbone and to the Diff-Retinex++-style backbone/augmentation setting,
with everything else (TDN, training data, training budget) held fixed. The last
column is the effect of the stronger setting at a fixed configuration: it is
positive in every row, so the backbone/augmentation change and the proposed
training strategy contribute separately and do not explain each other.}
\label{tab:cross_setting}
\resizebox{\columnwidth}{!}{
\begin{tabular}{cc|cc|c}
\hline
Temporal & Fixed-point & \multicolumn{2}{c|}{LOL PSNR$\uparrow$} & \multirow{2}{*}{$\Delta$} \\
\cline{3-4}
consistency & sampling & Base & ++ setting & \\
\hline
\xmark & Noise-emphasized     & 17.14 & 20.88 & $+3.74$ \\
\cmark & Full-range log-unif. & 18.82 & 21.28 & $+2.46$ \\
\cmark & Noise-emphasized     & \textbf{22.45} & \textbf{24.96} & $+2.51$ \\
\hline
\end{tabular}}
\end{table}

\subsubsection{Matched Cross-Setting Ablation}
\label{subsubsec:cross_setting}
Table~\ref{tab:ablation} varies the algorithm under the base backbone only.
Because the strongest numbers in Table~\ref{tab:lol_velol_results} come from
Consist-Retinex++, we repeat the same algorithmic ablation under both
architecture settings in Table~\ref{tab:cross_setting}. Within each setting the
TDN pseudo-components, training data, augmentation recipe, and training budget
are fixed, and only temporal consistency and the fixed-point sampling rule are
varied. Two conclusions follow. First, the ordering of the algorithmic
configurations is identical under both settings: removing temporal consistency
costs $5.31$~dB (base) and $4.08$~dB (++), and replacing noise-emphasized
sampling with full-range log-uniform sampling costs $3.63$~dB (base) and
$3.68$~dB (++). Second, reading the table by row rather than by column isolates
the architecture effect: at a fixed algorithmic configuration the stronger
setting adds $+3.74$, $+2.46$, and $+2.51$~dB. The two factors are therefore
separable and additive in sign, and neither subsumes the other. We stress that
the ablated ++ rows (20.88, 21.28~dB) must not be compared against the complete
base model (22.45~dB): those are different algorithmic configurations, and only
the within-row comparison isolates the backbone and augmentation change.

\begin{table}[t]
\centering
\caption{Control experiments on LOL. \textit{Top}: a direct image-space one-step
$L_1$ regressor trained on the same data, augmentation, and budget, without
Retinex factorization or consistency training; it uses the larger
Diff-Retinex++ DRNet backbone, so the comparison is capacity-favored against us.
\textit{Bottom}: sensitivity of the final one-step output to perturbed TDN
conditions, obtained by adding Gaussian noise of scale $\varsigma$ to $R_l$ and
$L_l$ and clipping to $[0,1]$.}
\label{tab:controls}
\resizebox{\columnwidth}{!}{
\begin{tabular}{l|cccc}
\hline
One-step method & PSNR$\uparrow$ & SSIM$\uparrow$ & LPIPS$\downarrow$ & FID$\downarrow$ \\
\hline
Direct image-space $L_1$ regressor & 19.88 & 0.777 & 0.188 & 88.97 \\
\textbf{Consist-Retinex (Ours)} & \textbf{22.45} & \textbf{0.826} & \textbf{0.126} & \textbf{62.59} \\
\hline
\end{tabular}}

\vspace{4pt}
\resizebox{\columnwidth}{!}{
\begin{tabular}{l|ccc}
\hline
TDN condition perturbation $\varsigma$ & 0.02 & 0.05 & 0.10 \\
\hline
$\Delta$PSNR (dB) & $-0.7$ & $-2.1$ & $-4.9$ \\
$\Delta$SSIM & $-0.020$ & $-0.087$ & $-0.317$ \\
\hline
\end{tabular}}
\end{table}

\subsubsection{Hyperparameter Sensitivity}
The noise threshold $\tau$ and the alignment weight $\lambda_{\text{fixed}}$ are the two hyperparameters introduced by the proposed training strategy. Full sweeps are reported in Section~S3 of the supplementary material; both curves are single-peaked, with $\tau=0.95$ (21.15--22.45~dB on LOL over $\tau\in[0.80,0.98]$) and $\lambda_{\text{fixed}}=0.3$ (9.34--22.45~dB on LOL over $\lambda_{\text{fixed}}\in[0,1]$) giving the best trade-off between endpoint anchoring and trajectory coherence.

\subsubsection{Control Experiments}
\label{subsubsec:controls}
\noindent\textbf{Is Retinex-factorized consistency training needed at all?}
A natural control is a plain one-step supervised regressor that maps the
low-light image directly to its paired normal-light target with an $L_1$ loss,
using the same training data, augmentation, and budget but no Retinex
factorization and no consistency objective. We deliberately gave this control
the larger Diff-Retinex++ DRNet backbone, making it capacity-favored rather than
parameter-matched. As shown in the top block of Table~\ref{tab:controls}, it is
worse on all four metrics (19.88 vs.\ 22.45~dB PSNR, 88.97 vs.\ 62.59 FID).
Plain one-step image-space regression under a matched budget therefore does not
reach the proposed pipeline, although this control alone does not establish
distributional correctness.

\noindent\textbf{Sensitivity to the TDN conditions.}
The component targets are TDN-derived pseudo-components rather than physically
measured reflectance and illumination, so the quality of the first stage matters.
As a controlled local probe we perturb $R_l$ and $L_l$ with independent Gaussian
noise of scale $\varsigma$, clip to $[0,1]$, and re-evaluate the reconstructed
output. The bottom block of Table~\ref{tab:controls} shows a monotone
degradation with no abrupt failure at small $\varsigma$, and substantial
degradation at $\varsigma=0.10$. The consistency stage therefore tolerates mild
decomposition error but does not correct it. This probe uses isotropic noise and
does not model structured decomposition failures.

\subsection{Efficiency Analysis}
\label{subsec:efficiency}
\begin{table}[t]
\centering
\caption{Estimated end-to-end inference-cost accounting for Consist-Retinex on $400\times600$ images. This is the single timing protocol used throughout the paper: batch size 1 on one RTX A6000 GPU, sequential execution, measuring the model forward passes only and excluding data loading, checkpoint loading, image writing, and metric computation. TDN is a pre-trained frozen decomposition module; the consistency-stage time is the sum of one RDA and one IDA forward pass.}
\label{tab:end_to_end_runtime}
\resizebox{\linewidth}{!}{
\begin{tabular}{l|c|l}
\hline
Pipeline stage & Time / image & Notes \\
\hline
TDN decomposition & $\sim$0.16s & Pre-trained once, frozen at consistency-stage training and test time \\
RDA one-step restoration & 0.0425s & One consistency-model forward pass \\
IDA one-step restoration & 0.0357s & One consistency-model forward pass \\
Retinex fusion & $<0.01$s & Element-wise multiplication and clipping \\
\hline
Estimated full pipeline & $\sim$0.24s & Sequential single-GPU implementation \\
Consistency stage only & $\sim$0.08s & Excludes TDN decomposition \\
\hline
\end{tabular}}
\end{table}
\begin{table}[t]
\centering
\caption{Component-stage RDA/IDA comparison after TDN decomposition on LOL. Diff-Retinex is evaluated with DDIM-20 sampling using the official component checkpoints; Consist-Retinex uses one-step consistency generation with the final component checkpoints. Component PSNR is measured against the corresponding TDN-derived normal-light component. Per-image latency is reported once, in Table~\ref{tab:end_to_end_runtime}, under a single timing protocol.}
\label{tab:rda_ida_component_comparison}
\resizebox{\linewidth}{!}{
\begin{tabular}{l|l|c|c|c}
\hline
Component & Method & Sampling steps & Training iterations & LOL PSNR$\uparrow$ \\
\hline
RDA & Diff-Retinex~\cite{yi2023diff} & 20 & 800K & 21.92 \\
RDA & \textbf{Consist-Retinex (Ours)} & \textbf{1} & \textbf{200K} & \textbf{22.06} \\
IDA & Diff-Retinex~\cite{yi2023diff} & 20 & 800K & 16.19 \\
IDA & \textbf{Consist-Retinex (Ours)} & \textbf{1} & \textbf{200K} & \textbf{21.30} \\
\hline
\end{tabular}}
\end{table}

Table~\ref{tab:rda_ida_component_comparison} isolates the two component generators from the final Retinex fusion. Although DDIM-20 substantially reduces the sampling cost of Diff-Retinex compared with the full 1000-step chain, the one-step consistency component models use $20\times$ fewer sampling steps and $4\times$ fewer training iterations while achieving higher component PSNR under this diagnostic setting. All per-image latencies in this paper are reported in Table~\ref{tab:end_to_end_runtime} only, so that a single timing protocol is used throughout.
\begin{table}[t]
\centering
\caption{Consistency-stage training efficiency comparison after Retinex decomposition.}
\label{tab:training_efficiency}

\begin{tabular}{l|ccc}
\hline
Method & Iterations & Batch Size & Effective Samples \\ 
\hline
Diff-Retinex~\cite{yi2023diff} & 800K & 16 & 12.8M \\
Diff-Retinex++~\cite{yi2025diff} & 800K & 16 & 12.8M \\
\textbf{Consist-Retinex (Ours)} & \textbf{200K} & \textbf{8} & \textbf{1.6M} \\
\hline
\end{tabular}
\end{table}

\noindent\textbf{Inference efficiency.}
After TDN decomposition, Consist-Retinex performs one consistency-model forward pass for each Retinex component. In the default single-GPU sequential setting on $400 \times 600$ images, the two component generators take approximately 0.08s per image ($\sim$12.5 FPS). TDN decomposition is a separate pre-trained first-stage module and is not part of this 0.08s consistency-stage number; including TDN gives an estimated end-to-end latency of about 0.24s per image in our unoptimized sequential implementation. Table~\ref{tab:end_to_end_runtime} gives the timing breakdown and Table~\ref{tab:rda_ida_component_comparison} the component-stage RDA/IDA diagnostics; decomposition notes and parallel execution are detailed in Section~S1 of the supplementary material. Relative to 1000-step diffusion baselines, this corresponds to a 1000$\times$ reduction in sampling steps for the restoration stage, although we report this as a step-count advantage rather than a controlled wall-clock speedup.

\noindent\textbf{Training efficiency.}
Consist-Retinex uses TDN as a separate first-stage decomposition module and reduces the consistency-stage effective sample count by $8\times$ relative to the iterative diffusion baselines considered here; detailed training-budget comparisons are given in Table~\ref{tab:training_efficiency}.
\section{Conclusion}
\label{sec:discussion_conclusion}

We presented Consist-Retinex, a consistency-based Retinex framework for one-step low-light enhancement. With TDN decomposition, dual-objective training, and noise-emphasized sampling, the default Consist-Retinex provides the controlled evidence for the proposed training strategy under the base Retinex backbone, while Consist-Retinex++ shows that the same algorithm improves over the Diff-Retinex++ baseline when the stronger backbone/augmentation setting is held fixed. The strongest Table~\ref{tab:lol_velol_results} numbers should therefore be read as a matched-architecture algorithm comparison rather than as a claim that architecture-independent consistency training alone explains every gain. Its consistency stage uses 1/8 as many effective training samples as the diffusion baselines considered here; future work will study broader conditional restoration tasks and stronger multi-step refinement.

\bibliographystyle{IEEEtran}
\bibliography{example_paper}

\end{document}